\documentclass[twoside,11pt]{article}

\usepackage[caption=false,font=footnotesize]{subfig}
\usepackage{jmlr2e}
\usepackage{graphicx}
\usepackage{amssymb,amsmath,bm}
\usepackage{textcomp}
\usepackage{enumitem}
\usepackage[margin=1.2in]{geometry}
\usepackage[export]{adjustbox}
\usepackage{multirow}

\newcommand{\mm}{\boldsymbol{\mu}}
\newcommand{\omg}{\boldsymbol{\omega}}
\newcommand{\Si}{\boldsymbol{\Sigma}}
\def\x{{\mathbf x}}
\def\T{{\mathbf T}}

\newcommand{\si}{{s_i}}
\newcommand{\ci}{{c_i}}
\newcommand{\mi}{{d_i}}

\jmlrheading{1}{2000}{1-48}{4/00}{10/00}{meila00a}{Marina Meil\u{a} and Michael I. Jordan}

\ShortHeadings{Joint PLDA}{Ferrer and McLaren}
\firstpageno{1}

\begin{document}

\title{Joint PLDA for Simultaneous Modeling of Two Factors}

\author{\name Luciana Ferrer \email lferrer@dc.uba.ar \\
       \addr Instituto de Investigaci\'on en Ciencias de la Computaci\'on (ICC)\\
 	   CONICET-Universidad de Buenos Aires\\
	   Buenos Aires, Argentina
       \AND
       \name Mitchell McLaren \email mitchell.mclaren@sri.com \\
       \addr Speech Technology and Research Lab (StarLab)\\
       SRI International\\
       Menlo Park, USA}

\editor{Kevin Murphy and Bernhard Sch{\"o}lkopf}

\maketitle

\begin{abstract}
Probabilistic linear discriminant analysis (PLDA) is a method used for biometric problems like speaker or face recognition that models the variability of the samples using two latent variables, one that depends on the class of the sample and another one that is assumed independent across samples and models the within-class variability. In this work, we propose a generalization of PLDA that enables joint modeling of two sample-dependent factors: the class of interest and a nuisance condition. The approach does not change the basic form of PLDA but rather modifies the training procedure to consider the dependency across samples of the latent variable that models within-class variability. While the identity of the nuisance condition is needed during training, it is not needed during testing since we propose a scoring procedure that marginalizes over the corresponding latent variable.   
We show results on a multilingual speaker-verification task, where the language spoken is considered a nuisance condition. We show that the proposed joint PLDA approach leads to significant performance gains in this task for two different datasets, in particular when the training data contains mostly or only monolingual speakers.
\end{abstract}


\section{Introduction}

PLDA \cite{prince:plda} was first proposed for doing inferences about the identity of a person from an image of their face. The technique was later widely adopted by the speaker recognition community, becoming the state-of-the-art scoring technique for this task \citep{kenny:10,burget:icassp11,em4plda,senoussaoui2011,matejka:11}. PLDA assumes that each sample is represented by a feature vector of fixed dimension and that this vector is given by a sum of three terms: a term that depends on the class of the sample, a term that models the within-class variability and is assumed independent across samples, and a final term that models any remaining variability and is also independent across samples. These assumptions imply that all samples from the same class are independent of each other and also independent of samples from other classes once the class is known.

In contrast with the assumptions made by PLDA, many training datasets consist of samples that come from a small set of distinct conditions. For example, many speaker recognition datasets contain only a few acoustic conditions (different microphones or noise conditions), speech styles (conversational, monologue, read), and languages. Samples corresponding to the same condition will most likely be statistically dependent.

The literature proposes a few approaches that generalize PLDA to consider metadata about the samples during training. The main motivation for these approaches, though, is not to relax the conditional independence assumption but rather to enable a more flexible model that can adapt to each of the available conditions rather than assuming that samples from all conditions can be modeled with the same linear model. Yet, a side effect of the proposed generalizations is the introduction of a dependency between samples from the same condition. The simplest approach of this family is to train a separate PLDA model for each condition, as proposed by \citet{garcia2012multicondition}. Nevertheless, in this paper, the authors show that pooling the data from all conditions, as proposed by \citet{lei:icassp12}, leads to better performance than training separate models. This result is reasonable, since training separate PLDA models does not allow the overall model to learn how samples from the same class vary across conditions; only within-condition variation is learned. 

The tied PLDA model proposed by \citet{peng2012} is designed to tackle this problem. In this approach, one PLDA model is trained for each condition, but these models are tied by forcing the latent variable corresponding to each class to be the same across all conditions. The approach was shown to outperform standard PLDA with pooled training data when each class in the training data is seen under both considered conditions, frontal and profile, in a face recognition task. A similar approach is proposed by \cite{mak2016mixture}, but in this case, the mixture component is not given during training. Instead, the PLDA mixture components depend on a continuous metadata value, which is modeled with a mixture of Gaussians. The approach is tested by adding noise to the training data at different SNR levels. The resulting training data then contains samples  for each speaker at different SNR levels. Under these conditions, the authors show gains from the proposed approach compared to pooling all the data to train a single PLDA model.

In this paper, we consider a scenario where each speaker in the training data is seen only under a small subset of the conditions present in the training set (potentially, only one). Further, we expect some conditions to have much less training data than others. Under this scenario, the tied PLDA approach does not work well, since it requires training a PLDA model of the same dimensions for each condition, which may be impossible or suboptimal for the conditions with less data. Further, the tied PLDA model can only learn how the nuisance conditions affect the classes of interest if it is provided samples for each class under different conditions during training. 

We propose a novel generalization of the PLDA model that relaxes the conditional independence assumption without increasing the size of the parameter space, keeping the same functional form of the original PLDA model but modifying the training and scoring procedures to consider the dependency across samples originating from the sample's condition. In the propose approach, which we call Joint PLDA (JPLDA), the condition is assumed to be known during training but not during testing. An expectation-maximization (EM) training procedure is formulated that takes into account the condition of each sample. Scoring is performed, as in standard PLDA, by computing a likelihood ratio between the null hypothesis that the two sides of a trial belong to the same speaker versus the alternative hypothesis that the two sides belong to different speakers. The two likelihoods are computed by marginalizing over two hypotheses about the condition in both sides of a trial: that they are the same and that they are different. This way, we expect that the new model will be better at coping with same-condition versus different-condition trials than standard PLDA, since knowledge about the condition is used during training and implicitly considered during scoring. Further, we expect this model to behave better than tied PLDA under a training scenario where the number of samples is highly imbalanced across conditions and each speaker is seen only under one or a small subset of conditions.

We show results on two multilingual speaker recognition datasets, one composed of Mixer data \citep{cieri:2007} from the speaker recognition evaluations organized by NIST and another that uses LASRS data \citep{Beck2004}. We evaluate two training scenarios, one using all available training data from the PRISM dataset \citep{ferrer:sre11}, which contains a small percentage of speakers speaking two different languages, and one where we subset the training data to contain only one language per speaker. We show that JPLDA significantly outperforms two standard PLDA approaches with different structures and tied PLDA, especially when the training data contains mostly or only a single language per speaker.

\section{Standard PLDA Models}

In this work, we adopt the nomenclature usually used by the speaker recognition community. Yet, the model proposed can be used for the original image processing task or any other task for which standard PLDA is used.

Standard PLDA \citep{prince:plda} assumes that the vector $m_i$ representing a certain sample from speaker $s_i$ is given by
\begin{equation}
m_i = \mu + V y_\si + U x_i + z_i,
\end{equation}
where $\mu$ is the global mean of the training data; $y_\si$ is a vector of size $R_y$, the dimension of the speaker subspace; and $x_i$ is a vector of size $R_x$, the dimension of the subspace corresponding to the nuisance condition or, as usually called in speaker recognition, the channel. The model assumes that
\begin{eqnarray}
y_\si & \sim & N(0,I), \\
x_i & \sim & N(0,I), \\
z_i & \sim & N(0, D^{-1}),
\end{eqnarray}
where the matrix $D$ is assumed to be diagonal. All these latent variables are assumed independent: speaker variables are independent across speakers, and the nuisance variable $x_i$ and noise variable $z_i$ are independent across samples. 

The model described above corresponds to the original PLDA formulation, which we will call full PLDA (FPLDA). In speaker recognition, a simplified version of PLDA (SPLDA for the purpose of this paper) is more commonly used, where the matrix $V$ is full rank, and the nuisance factor is absorbed into the noise factor, which is then assumed to have a full rather than diagonal covariance matrix. This simpler model was shown to give better performance than the original model in some publications. \citet{sizov2014} gives a comprehensive explanation of the usual flavors of PLDA. 

The training of PLDA parameters is done using an EM algorithm. The EM formulation for SPLDA and FPLDA can be found in two very detailed documents by \cite{em4plda,em4splda}. We will not reproduce the EM formulas here, but we will describe the two initialization procedures we use, since they will be compared in the experimental section.

\subsection{EM Initialization Procedure}
\label{sec:plda_init}

The EM algorithm requires an initial model to start the iterations. This model can be generated randomly or, in the case of SPLDA, with a ``smart'' procedure that results in a much better initial model that, in turn, requires many fewer or no EM iterations to converge to the final parameters.

In our experiments, for random initialization, we set $D$ to be an identity matrix, and $V$ and $U$, when applicable, to be matrices with random elements drawn from a normal distribution with standard deviation 0.01 and mean 0.

For SPLDA, we also try a smart initialization approach that is commonly used and well motivated \citep[see][for an explanation of why it is reasonable to use this initialization]{sizov2014}, given by
\begin{eqnarray}
V & = & Q \Lambda^{-1/2},  \\
D & = & W, 
\end{eqnarray}
where $W$ is the empirical within-class covariance matrix of the training data, and $Q$ is a matrix with the eigenvectors corresponding to the $R_y$ largest eigenvalues of the between-class covariance matrix of the training data and $\Lambda^{-1/2}$ is a diagonal matrix containing the square root of those eigenvalues.

\subsection{Scoring}

In this work, we consider a verification task. Two sets of samples, an enrollment set $E$ and a test set $T$, each corresponding to one or more samples from the same class, are compared to decide whether the two classes are the same or different. This comparison is usually called a \emph{trial} in speaker verification. In some applications, a hard decision is needed; in others, a soft score is preferable. The PLDA paper \citep{prince:plda} proposed to use the likelihood ratio (LR) between the two hypotheses as a score. This score can then be thresholded to make hard decisions if required. The LR is given by
\begin{eqnarray}
LR & = & \frac{p(E,T|H_{SS})}{p(E,T | H_{DS})},
\end{eqnarray}
where $H_{SS}$ is the hypothesis that the speakers in both sets are the same, while  $H_{DS}$ is the hypothesis that the speakers are different. This value can be computed using a closed form using the PLDA model. In our code we use the formulation derived by \citet{cumani2014}, Equation (34). Note, though, that the last term in that equation should not be there (this mistake was confirmed by one coauthor of the paper).  

\section{Tied PLDA Model}

The tied PLDA model was introduced by \citet{peng2012}. The model is a mixture of PLDA models where the latent variable corresponding to the speaker is tied across components. 
\begin{equation}
m_i = \mu_\mi + V_\mi y_\si + U_\mi x_i + z_i,
\end{equation}
where $\mi$ indicates the mixture component corresponding to sample $i$, $\mu_\mi$ is the mean of the data for component $\mi$, and
\begin{eqnarray}
y_\si & \sim & N(0,I), \\
x_i & \sim & N(0,I), \\
z_i & \sim & N(0, D_\mi^{-1}).
\end{eqnarray}
Hence, once the mixture component is given, the model reduces to a standard PLDA model. In this work, we assume that the mixture component is known both during training and during testing, as in the original work \citep{peng2012}, though the authors note that this is not a necessary condition. In the simplest case we could take the mixture component to be the nuisance condition of the sample but, as we will see, this might not be feasible if some conditions have too few training samples in which case grouping of samples from different conditions into the same component might be necessary. Note that the latent variable $y_\si$ does not depend on the component. Rather, this variable is tied for all samples from the same speaker across components. This enables the model to properly represent cross-component variability.

As for the original PLDA model, a simple PLDA model can be used instead of the full PLDA model for each component in the mixture. Further, the covariance matrix for the noise term can be either full or diagonal. In this work, each component is described by a SPLDA model for simplicity of implementation, since the difference between SPLDA and FPLDA is very small in practice.

The TPLDA model described by \citet{peng2012} and used here coincides with what \citet{mak2016mixture} calls SNR-dependent mixture PLDA model if we assume the SNR to be discrete rather than continuous so that the posterior probability for each component is fixed to 1 for the component corresponding to the sample, and to 0 otherwise. The training and scoring procedures for TPLDA can be found in the supplementary material for \citet{mak2016mixture}.

\section{Joint PLDA Model}

In this work, we propose a generalization of the original model where the nuisance variable is no longer considered independent across samples, but potentially shared (tied) across samples that correspond to the same nuisance condition. This makes the model symmetric in the two latent variables (corresponding to the class of interest and the nuisance condition) in the sense that both variables are tied across all samples sharing a certain label. To represent this dependency, we introduce a  condition label for each sample, called $\ci$. Given this label, and the speaker label $\si$, we propose to model vector $m_i$ of dimension $R_m$ for sample $i$ as:
\begin{equation}
m_i = \mu + V y_\si + U x_\ci + z_i,
\end{equation}
where, as before, $y_\si$ is a vector of size $R_y$ and $x_\ci$ is a vector of size $R_x$, and
\begin{eqnarray}
y_\si & \sim & N(0,I), \\
x_\ci & \sim & N(0,I), \\
z_i & \sim & N(0, D^{-1}).
\end{eqnarray}
The model's parameters to estimate are $\lambda = \{\mu, V, U, D\}$, as in the standard PLDA formulation, but the input data for the training algorithm is now required to have a second set of labels indicating the nuisance condition of each sample. 

The expectation-maximization equations for training this new model are significantly more involved than for the original PLDA model. This is due to the fact that each speaker cannot be treated separately from the others since samples from one speaker might be dependent on samples from a different speaker. This creates a potential dependency between all training samples, which greatly complicates the formulation, increasing the computation time by orders of magnitude for each EM iteration. Nevertheless, as we will see in the experiments, initializing the model in a smart way basically makes EM unnecessary in our experiments, reducing the training time of the model to just a small factor of what is required to train standard PLDA on the same data. A detailed derivation of the EM algorithm and scoring procedure for JPLDA is given by \citet{ferrer2017joint}. Here we only describe the initialization procedure used for training the model with EM and the form used for the LR.

The matrix $D$ in the JPLDA model can be full or diagonal. If we want $D$ to be diagonal, we simply set $D$ to be the diagonal part of the estimated value for $D$ in each maximization step of the EM algorithm, as done for the standard PLDA EM algorithm \citep{em4plda}.

\subsection{EM Initialization Procedure}

The JPLDA model can be randomly initialized using the same procedure as for standard PLDA described in Section \ref{sec:plda_init}. Note that, as for standard PLDA, $\mu$ is not iteratively estimated but set to the global mean of the training data.

We propose the following alternative procedure to get the initial values for the PLDA model, $U_0$, $V_0$ and $D_0$:
\begin{itemize}[noitemsep,topsep=0pt]

\item Estimate a ``condition'' SPLDA model using the original matrix $M$ (of size $NxR_m$, where $N$ is the number of training samples) of training vectors, and the nuisance conditions as labels instead of the speakers, setting the rank of $V$ to be the desired condition rank $R_x$. Call the $V$ matrix of this model $V_c$.

\item Estimate the matrix of latent variables from the condition SPLDA model for all training samples. This matrix $X$ has dimension $NXR_x$. Note that all samples with the same condition label will have the same latent variable.

\item Create new training vectors by subtracting the effect of the nuisance condition as follows: $M_c = M - X V^T$.

\item Estimate a ``speaker'' SPLDA model using $M_c$ as training data and the speakers as labels, setting the rank of $V$ to be the desired speaker rank $R_y$. Call the $V$ matrix of this model $V_s$ and the $D$ matrix of this model $D_s$.

\item Set $U_0=V_c$, $V_0=V_s$, and $D_0=D_s$. If we wish to produce a JPLDA with diagonal $D$, then we simply take $D_0$ to be a diagonal matrix with the same diagonal as $D_s$.

\end{itemize}

As we will see this ``smart'' initialization leads to such a good starting point that EM iterations are unnecessary in our experiments.

\subsection{Scoring}
\label{sec:jplda_scoring}

As for standard PLDA, we define the score as the likelihood ratio between the two hypotheses: that the speakers are the same and that the speakers are different. Nevertheless, in this case we need to marginalize both likelihoods over two new hypotheses: that the nuisance conditions are the same and that they are different. This is because, in general, we cannot assume that the nuisance condition is known during testing. Hence, the LR is computed as follows:
\begin{eqnarray}
LR & = & \frac{p(E,T|H_{SS},H_{SC})P(H_{SC}|H_{SS}) + p(E,T|H_{SS},H_{DC})P(H_{DC}|H_{SS})}{p(E,T|H_{DS},H_{SC})P(H_{SC}|H_{DS}) + p(E,T|H_{DS},H_{DC})P(H_{DC}|H_{DS})}
\end{eqnarray}
where, as before, $H_{SS}$ is the hypothesis that the speakers for both sets are the same, and  $H_{DS}$ is the hypothesis that they are different, while $H_{SC}$ is the hypothesis that the nuisance condition for both sets is the same, and  $H_{DC}$ is the hypothesis that they are different. This LR value can be computed using a closed form derived in \citep{ferrer2017joint}. 

Note that here we assume that all samples from the enrollment set come from the same condition and all samples from the test set come from the same condition, which could be the same or different from the enrollment condition. This is trivially true when the sets are composed of a single sample, which is the case we consider in the experiments in this paper. The formulation would become more complex without this assumption since we would need to consider the possibility that each sample in each set could come from different conditions.  In \citep{ferrer2017joint}, we also derive the scoring formula for a multi-enrollment single-test case where the enrollment conditions are known and different from the test condition. This formulation is used when applying JPLDA to language identification (LID). Experiments on LID will be the subject of another paper. Further, the generalization of the scoring formula to multiple enrollment or test samples with unknown conditions will be considered in future work.

The scoring formula above depends on two prior probabilities, the probability that the enrollment and test conditions are the same given that the speakers are the same, $P(H_{SC}|H_{SS})$, and the probability that the conditions are the same given that the speakers are different, $P(H_{SC}|H_{DS})$. The other two prior probabilities are dependent on these two since $P(H_{SC}|H_{SS})+P(H_{DC}|H_{SS})=1$ and $P(H_{SC}|H_{DS})+P(H_{DC}|H_{DS})=1$. These two independent prior probabilities are parameters that could be computed from the training data, tuned using a development set, or set to arbitrary values based on what is known about the test data.

In some applications, the nuisance condition might be known also in testing. In that case, the same-condition priors can be set to 1.0 for same-condition trials and to 0 for different-condition trials. 

\section{Experimental Setup}

In this section we describe the task, the performance metrics, the data used for the experiments and the procedure used to convert each audio sample to a fixed-length vector to be modeled by the different PLDA methods, as well as the method used to calibrate the scores for some of the results in Section \ref{res:train}.

\subsection{Multilanguage Speaker Verification}

The task considered for our experiments is speaker verification, which consists of determining whether two sets of samples, an enrollment and a test set, belong to the same speaker or not. Here we consider the simplest case, where both enrollment and test sets each contain a single speech sample. A pair of enrollment and test samples is called a \emph{trial}. A trial is a target trial if the enrollment and test speakers are the same and an impostor trial if the two speakers are different. In this paper we explore the problem of multilanguage speaker verification where test trials can be composed of two samples in the same language (same-language trials) or two samples in different languages (cross-language trials). 

Most state-of-the-art speaker verification systems are inherently language-independent in the sense that they do not use information about the language spoken in order to generate the output score. Yet, this does not mean that they are robust to language variation. In fact, speaker verification performance is known to degrade significantly in cross-language trials as well as in same-language trials from languages not found in the training set \citep{auckenthaler2001,misra2014,rozi2016}. 

\citet{rozi2016} discusses a problem that occurs when training PLDA models with multilingual data: the distribution of the speaker factors becomes broader to cover the different languages that a speaker might speak, which could result in suboptimal performance on same-language trials. They propose to mitigate this problem by training a standard PLDA model using both language and speaker as targets (i.e., samples from the same speaker but different language are considered as different speakers). This language-aware PLDA model performs significantly better on same-language trials than the model trained with speaker targets, but degrades on cross-language trials, since it cannot model cross-language variation. JPLDA, on the other hand, can simultaneously model language and speaker factors, allowing the speaker factors to keep a sharper distribution, while still modeling the effect of language, resulting in improved performance both in same-language and cross-language trials with respect to standard PLDA.

\subsection{Performance Metrics}

We compute performance using three different metrics: the equal-error rate (EER), the cost of likelihood ratio (Cllr), and detection error (DET) curves. The DET curves and the EER measure the performance of a system that uses the scores (in our case, the LRs) to make final decisions on the label of each sample by comparing these scores with a ``decision threshold.'' Samples whose scores are above the threshold are labelled as targets and samples whose scores are below the threshold are labelled as impostors. Two types of error are then possible: (1) misses, the true target trials that are labelled as impostors by the system, and (2) false alarms, the impostor trials that are labelled as targets by the system. The EER is given by the miss rate when the decision threshold is set such that the miss rate is equal to the false alarm rate. 

DET curves \citep{martin1997det} are a variation over the traditional receiver operating characteristic (ROC) curves that have been widely used for speaker verification for two decades. As with the ROC curves, DET curves show the performance of the system over a range of decision thresholds rather than focusing on a single point, as the EER does. 
A DET curve is a plot of the false alarm rate versus the miss rate obtained while sweeping a decision threshold over a certain range where the axes are transformed to a probit scale. The probit transformation, the inverse of the cumulative distribution function of the standard normal distribution, converts the miss versus false alarm rate curve into a straight line if the score distribution for the two classes is Gaussian with the same standard deviation \citep{martin1997det}, which is a reasonable approximation for many speaker verification systems. 

Both EER and DET curves measure performance in a way that is insensitive to monotonic transformations of the scores. For this reason, they do not measure the quality of the scores themselves as likelihood ratios but only the discriminative power in these scores when used for decision making by thresholding. \citet{Brummer:csl06} proposed a way to measure the quality of the scores as true likelihood ratios for the speaker verification task using a logarithmic cost function that they call Cllr. Cllr measures both the discrimination and calibration of the system, where calibration refers to how close the scores are to true likelihood ratios for the task. Further, it measures the quality of the scores without committing to any specific operating point or decision threshold. Cllr has been widely used for evaluation of speaker verification systems for more than a decade. In this paper, we show Cllr performance for the final set of results as a complement to EER and DET curves. 


\subsection{Training Data}
\label{sec:training_data}

We consider two training conditions, one that includes all our available training data (FULL) and a subset that keeps only one language for each speaker (SINGLE-LAN). The second condition is designed to help us analyze performance of the PLDA methods under this extremely challenging scenario where no explicit information is available in the training data of the effect that language has on the vectors representing the samples.

The FULL training set is composed of:
\begin{itemize}[noitemsep,topsep=0pt]
\item Switchboard Cellular Part 1 \citep{swcellp1} and Cellular Part 2 \citep{swcellp2}, consisting of English cellphone conversations 
\item Switchboard 2 Phase 2 \citep{swph2} and Phase 3 \citep{swph3} samples, consisting of English telephone conversations 
\item Mixer data \citep{cieri:2007} from the 2004 to 2008 speaker recognition evaluations organized by the National Institute of Standards and Technology (NIST). This data contains English samples recorded both on telephone and microphone channels and non-English samples recorded on telephone channels. With very few exceptions,  speakers that recorded non-English samples also recorded English samples. Only one speaker has data in two non-English languages and no data in English. Only a subset of this data is used for training, leaving some speakers out for testing (Section \ref{sec:test}). We also discard data from languages for which only one or two speakers are available and samples where the language was unavailable or ambiguous (e.g., more than one language listed in the language key) in NIST's keys. 
\end{itemize}

The SINGLE-LAN training set is created by randomly keeping the samples from only one of the languages spoken by each speaker from the FULL training set.

Finally, in Section \ref{res:train}, we analyze results when subsetting these two training sets to contain a more balanced representation of channels while keeping all data from bilingual speakers for the FULL training set. Specifically, we subset the FULL training set to discard all the telephone data from speakers that do not have data in both English and another language. This is data that is not adding much new information, since all non-English data is recorded over telephone line, and all speakers with non-English data also have telephone recordings in English. By subsetting the data this way, we achieve a more balanced representation of the telephone data with respect to the microphone data, while emphasizing the data from bilingual speakers, which is a very small minority on the original set including all the data. To create the subset for the SINGLE-LAN training set, we simply keep the samples that appear in the subset from the FULL training set and also appear in the SINGLE-LAN set.

Table \ref{table:train} shows statistics on the two training sets and their corresponding subsets. The languages included under ``Other'' are: Arabic (with 440 samples); Bengali (88); French (25); Chinese (868); Farsi (25); Hindi (143); Italian (11); Japanese (124); Korean (78); Russian (478); Spanish (170); Tagalog (26); Thai (185); Vietnamese (169); Chinese Wu (63); and Cantonese (216).

\begin{table}[!t]
\renewcommand{\arraystretch}{1.3}
\caption{Statistics for the four training sets considered in our experiments. ``Eng'' refers to English data while ``Other'' refers to any language other than English. ``Phn'' refers to telephone or cellphone data, and ``Mic'' refers to all other microphones in the training set. ``Monoling'' refers to speakers for which we only have samples in a single language, and ``Biling'' refers to speakers for which we have samples in two languages (English plus one other language in most cases).}
\label{table:train}
\centering
\begin{tabular}{|c|c|c|c|c|c|c|c|c|c|}
\hline
& & \multicolumn{4}{c|}{Sample count} & \multicolumn{4}{c|}{Speaker count} \\ \cline{3-10}
Name & Sel & Eng & Eng & Other & Total & \multicolumn{2}{c|}{MonoLing} & BiLing & Total \\ \cline{7-8}
& & Mic & Phn & Phn     &       & Eng      & Other &  &  \\
\hline
\multirow{2}{*}{FULL} & all     &  11017 & 38382 & 3109 & 52508 & 2764 & 34 & 495 & 3293 \\
                     & subset  &  11017 & 3711  & 2733 & 17461 & 207 & 0 & 494 &  701 \\
\hline
\multirow{2}{*}{SINGLE-LAN} & all    & 10797 & 36619 & 1688 & 49104 & 3026 & 267 & 0 & 3293 \\
                            & subset & 10797 & 1948  & 1332 & 14077 & 468  & 233 & 0 & 701 \\
\hline
\end{tabular}
\end{table}

\subsection{Test Data}
\label{sec:test}

We consider two testing conditions, one composed of Mixer data and used for development and one composed of LASRS data and used as held-out set for final evaluation of the selected methods.

{\bf The Mixer test data} is composed of telephone samples from Mixer collections \citep{cieri:2007} from the 2005 to 2010 NIST speaker recognition evaluations, from speakers not used for training. We include 119 samples in Arabic from 21 speakers; 200 samples in Russian from 47 speakers; 309 samples in Thai from 38 speakers; 827 samples in Chinese from 163 speakers; and 5755 samples in English from 701 speakers (including those that also speak one of the other languages). 

The trials are created by selecting the same number of target and impostor same-language and cross-language trials such that the final set of trials is a balanced union of both types of trials. Further, the same-language trials are created as a balanced union of English versus non-English trials. The final set of trials contains 11,522 target trials and 858,119 impostor trials. 

{\bf The LASRS test data} is composed of samples from a bilingual, multi-model voice corpus \citep{Beck2004}. The corpus is composed of 100 bilingual speakers from each of three languages: Arabic, Korean and Spanish. Each speaker is asked to perform a series of tasks in English and also in their native language. Each task is recorded using several recording devices and repeated in two separate sessions recorded on different days. For our experiments, we use the conversational data for the Korean and Spanish speakers (we were unable to obtain the Arabic data). The LASRS trials are created by enrolling with data from the first recorded session and testing on the second recorded session in each of the two spoken languages. This results in approximately the same number of same-language and cross-language trials for a total of 692 target trials and 79484 impostor trials for each of seven different microphones: a camcorder microphone (Cm); a Desktop microphone (Dm); a studio microphone (Sm); an omnidirectional microphone (Om); a local telephone microphone (Tm); a remote telephone microphone (Tk); and a telephone earpiece (Ts). For this study, we only consider same-microphone trials for simplicity of analysis. For more details on the collection protocol, see \citep{Beck2004}.

\subsection{I-vector Extraction}

For validation of the proposed approach, we use a traditional i-vector framework for speaker recognition \citep{Dehak11}. I-vectors are finite-length vectors that represents the contents, or the total variability, of the speech in an audio recording. They are extracted using factor analysis with a single factor to describe all the variability in the observed features. Given a set of short-term features for the speech signal, $x = \{x_1, \ldots, x_T\}$, the i-vector model assumes each of these vectors is produced by the following model:
\begin{equation}
\x_t \sim \sum_k \gamma_{kt}\mathcal{N}(\mm_k + \T_k \omg , \Si_k)
\end{equation}
where the indices $k$ are the components of a Gaussian mixture model; the $\T_k$ matrices describe a low-rank subspace (called the total variability subspace) by which the means of the Gaussians are adapted to a particular speech segment; $\omg$ is a segment-specific, normal-distributed latent vector; $\mm_k$ and $\Si_k$ are the mean and covariance of the unadapted $k$-th Gaussian; and $\gamma_{kt}$ encodes the soft assignment of the features at time $t$, $x_t$, to class $k$. We compute the assignments as the posterior of the $k$-th Gaussian (before adaptation to the sample), given the features. The i-vector used to represent the speech signal is the maximum a posteriori (MAP) point estimate of the latent vector $\omg$.

In our experiments, the process for extracting an i-vector to represent a variable-length speech recording is as follows:
\begin{itemize}[noitemsep,topsep=0pt]
\item The first 20 mel-frequency cepstral coefficients (MFCCs) are extracted from the audio signal using a 25ms window every 10ms. MFCCs are an acoustic feature vector that captures information regarding the amplitude of different frequencies in a similar manner to how sounds are perceived by the human ear \citep{davis1980comparison}. The MFCCs are appended with deltas and double deltas to help capture the dynamics of speech over time \citep[e.g.,][]{gales2008application}. This results in a feature vector of 60 dimensions, with 100 frames (of feature vector) per second.
\item Speech activity detection (SAD) is applied to remove any frames that do not contain speech. For this purpose we use a deep neural network (DNN)-based model trained on telephone and microphone data from a combination of Fisher \citep{fisher1,fisher2}, Switchboard \citep{swcellp1,swcellp2,swph2,swph3} and Mixer data \cite{cieri:2007}, as well as a 30-minute long dual-tone multi-frequency (DTMF) signal without speech, and a set of 3740 signals where speech from the Fisher corpora was corrupted with non-vocal music at different SNR levels. We use MFCC features, mean and variance normalized using a sliding window of two seconds, and concatenated over a window of 31 frames. The resulting 620-dimensional feature vector forms the input to a DNN that consists of two hidden layers of sizes 500 and 100. The output layer of the DNN consists of two nodes trained to predict the posteriors for the speech and non-speech classes. These posteriors are converted into likelihood ratios using Bayes rule (assuming a prior of 0.5), and a threshold of 0.5 is applied to obtain the final speech regions.
\item Using the speech frames from all the training data, a Gaussian mixture model of 2048 components called the universal background model (UBM) is estimated. This UBM defines the parameters $\mm_k$ and $\Si_k$ of the i-vector extractor, as well as the prior probability of each component. 
The training data used for this estimation was given by a random subset of 10,000 samples of the data used to estimate the $T_k$ matrices described next. 
\item The subspaces $T_k$ are then estimated using the FULL training data described in Section \ref{sec:training_data}, except that samples from languages for which only one or two speakers are available or where the language was unavailable or ambiguous are not discarded for this purpose.
\item Finally, once the model is trained, the i-vectors for any audio sample can be extracted using only the speech frames, as in training. 
\end{itemize}

The i-vectors can then be used for determining speaker similarity between two utterances using PLDA. For the experiments, we process the i-vectors with multiclass linear discriminant analysis (LDA) trained on the same training data used for PLDA, after which we subtract the mean over the training data and perform length normalization\citep{romero:lennorm}. The length-normalization step serves to better satisfy the Gaussianity assumption behind PLDA. 

\subsection{Calibration}

Despite the fact that PLDA is designed to compute likelihood ratios, these scores are well known to be generally not well-calibrated (i.e., they are not true likelihood ratios for the task), probably due to inaccuracies in the assumptions made by the model. For this reason, to compute reasonable values for the Cllr, we need to first calibrate our systems. The optimization of score calibration is a complex issue that has been widely studied in the literature \citep[see, for example,][]{Brummer:csl06,brummer2014comparison,mclaren2014trial,mclaren:is2016calibration} and is out of the scope of this paper. Here we use a simple calibration approach that is widely used in the speaker verification literature where scores are transformed with a linear function whose parameters are trained with a logistic regression objective \citep{brummer2013likelihood}. In our experiments, a separate transformation is learned for each test condition using cross-validation.

\section{Results}

In this section we compare results for different EM initialization techniques, parameter settings and training data for the proposed and the baseline PLDA techniques described in previous sections.

The nuisance condition for JPLDA in these experiments is the language spoken in the sample. During training, this label is known; during scoring, the label is marginalized to compute the LR, unless otherwise indicated. For TPLDA, on the other hand, we cannot take the mixture component to be the language spoken in the sample. This is because we do not have enough training speakers for each language to train a good PLDA model for each component. Hence, we consider a two-component model with a component modeling all English data and another component modeling the non-English data. This, as we will see, turns out to be a good model when matched data is available for training both components. Note that our implementation of TPLDA assumes that the mixture component is given both in training and in scoring. This is possible in our experiments because we have the language spoken during testing. 

For SPLDA, FPLDA and JPLDA, the LDA dimension is set to 400; no dimensionality reduction is done in these cases but the data is still transformed by the LDA matrix, centered and length normalized. For TPLDA, on the other hand, we use an LDA dimension of 200, because we found that this value gives significantly better performance than keeping the original dimension of 400.

The speaker and language ranks for  all experiments in this section are fixed to 200 and 16, respectively. These values were chosen for being optimal or approximately optimal for all methods under study (FPLDA, SPLDA and JPLDA) when using all available training data. The speaker rank of 200 was chosen for being optimal or very close to optimal for all methods. The language rank of 16 is the largest rank that can be used for JPLDA. This value turned out to be optimal for JPLDA. FPLDA is largely insensitive to this parameter, giving very similar performance for language ranks between 5 and 16. Unless otherwise stated, all JPLDA results are obtained using $P(H_{SC}|H_{SS})=P(H_{SC}|H_{DS})=0.5$. For TPLDA we use a diagonal matrix for the covariance of the noise model which proved to be slightly better than a full covariance. 

All tuning decisions above were made based solely on the results on Mixer test data.

\subsection{Initialization and Convergence of Training Procedure}

We first show results for SPLDA, FPLDA and JPLDA as a function of the number of EM iterations run for the two initialization procedures, random and smart, explained in Section \ref{sec:plda_init}. For FPLDA, no standard way exists of which we are aware to smartly initialize all parameters of the model. In this case, we only show results for random initialization. For this section, we use the FULL training data without subsetting and test on the Mixer data.

\begin{figure}[!t]
\centering
\includegraphics[width=0.6\columnwidth]{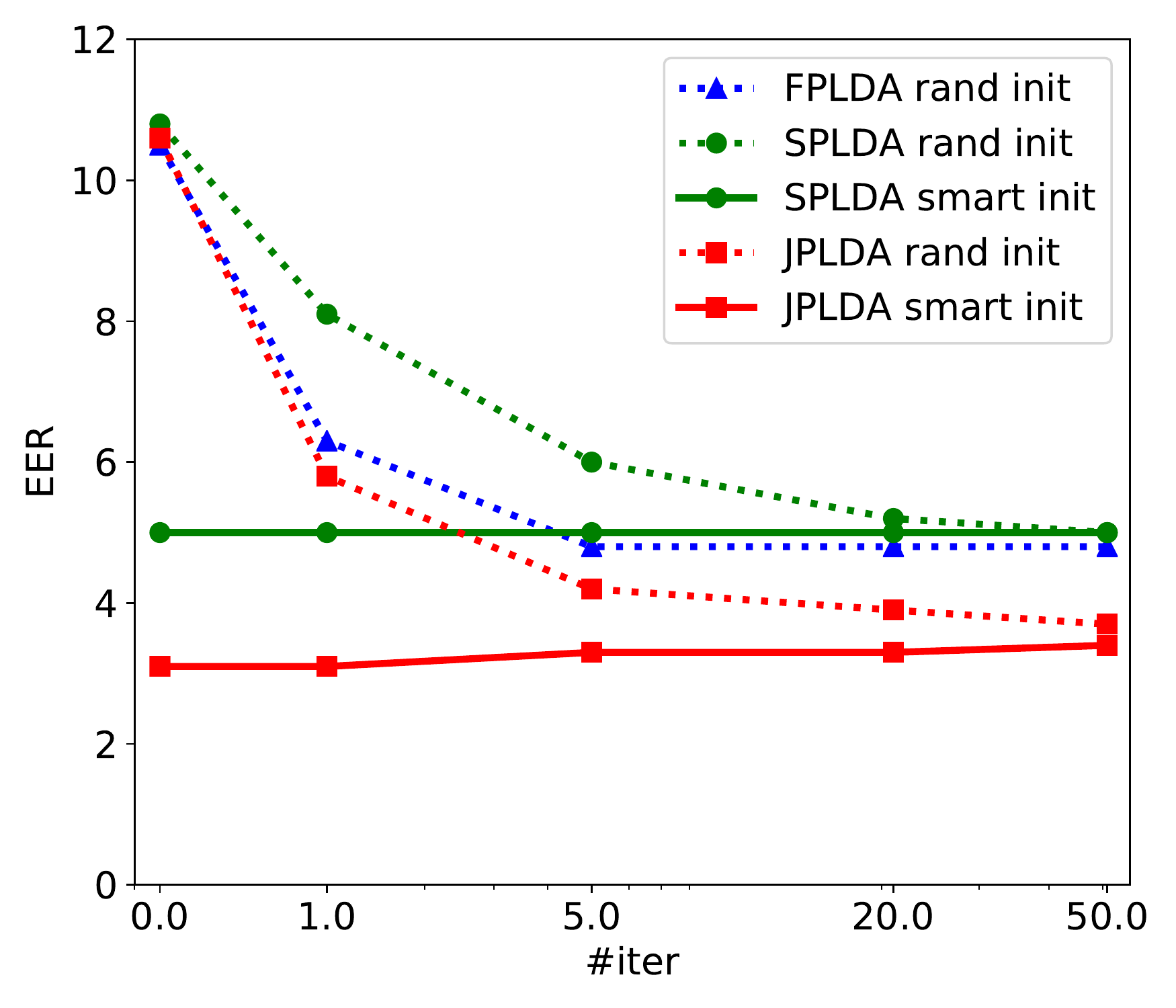}
\caption{Comparison of performance as a function of the number of EM iterations on the Mixer  test data using all available training data for random and smart initialization for three different PLDA models. Note that a log scale is used for the x-axis.}
\label{fig:iter}
\end{figure}

Results in Figure \ref{fig:iter} show that EM iterations are essential when random initialization is used, leading to large gains over the initial random model as the iterations progress and converging to an approximately stable value when reaching 50 iterations. On the other hand, when smart initialization of SPLDA or JPLDA is used, EM iterations are not necessary on thiss dataset. In fact, JPLDA performance with smart initialization slightly degrades for larger number of iterations, probably due to overfitting of the training data. For this reason, for the rest of the experiments we use only one iteration of EM for JPLDA, though zero iterations could also be safely used.

\subsection{Prior probability of same language in JPLDA}

In this section we show JPLDA results on the Mixer development set when using all available training data as a function of the prior probabilities of same language, $P(H_{SC}|H_{SS})$ and $P(H_{SC}|H_{DS})$ (see Section \ref{sec:jplda_scoring}). We fix these two parameters to the same value and sweep this value between 0 and 1 at 0.1 steps. We show results on the full test data but also split the data into same-language and cross-language trials. We compare these results with those we would obtain by knowing the language of each sample a priori and using this knowledge during scoring to set the priors appropriately as explained in Section \ref{sec:jplda_scoring}.

\begin{figure}[!t]
\centering
\includegraphics[width=0.6\columnwidth]{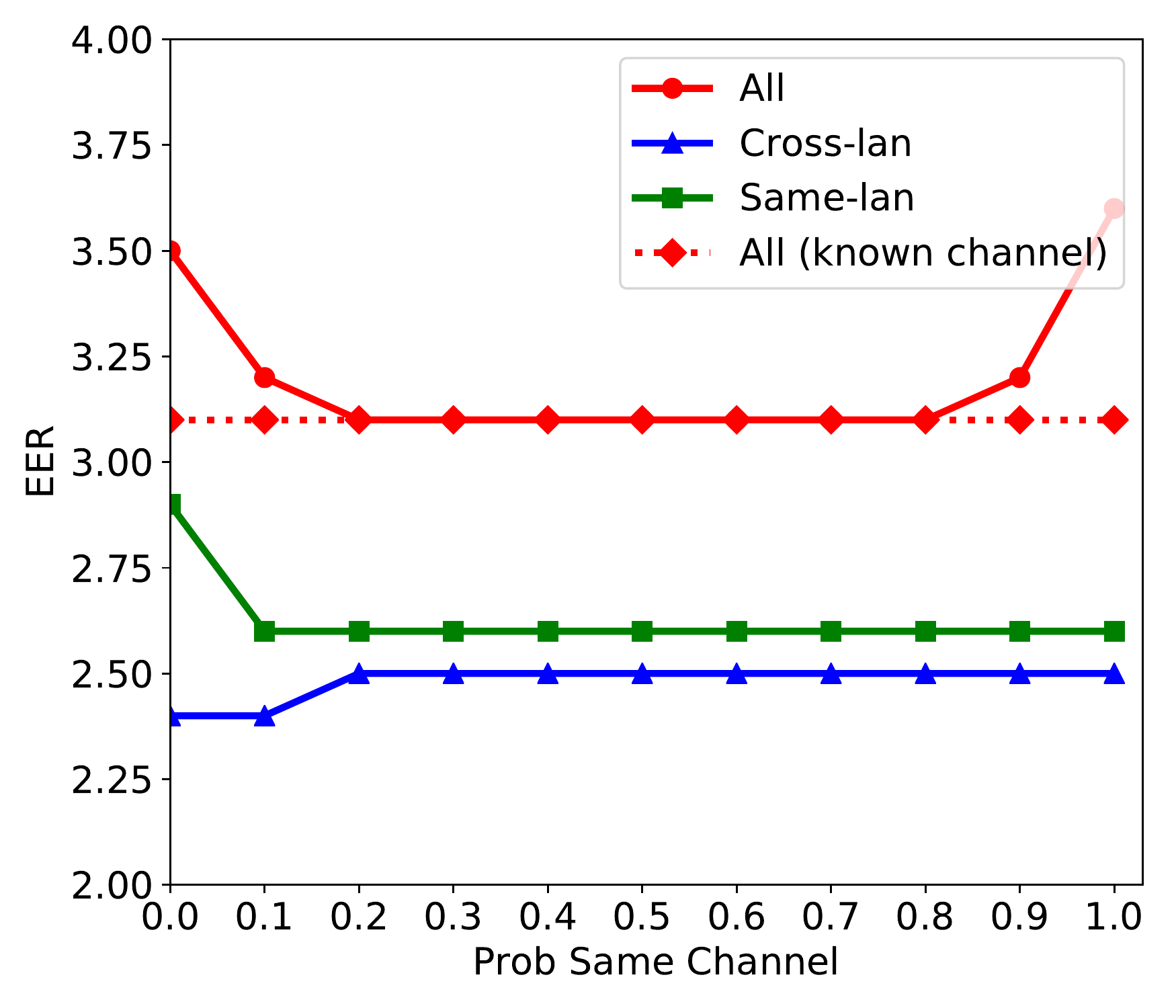}
\caption{Comparison of JPLDA performance as a function of the prior probability of same language on the Mixer test data using the FULL training data. The known-channel line corresponds to the performance on all trials when using the information about the test language during scoring. }
\label{fig:psc}
\end{figure}

Figure \ref{fig:psc} shows performance as a function of the probability of same language parameter. Values below 0.1 are optimal for the cross-language trials, while values above 0.1 are optimal for same-language trials. 
Once all trials are pooled together, values between 0.2 and 0.8 give almost identical performance. For this range of values, we can also see that performance is the same as what we would obtain if the language of the test files was known during scoring (the red dashed line in the plot). This performance is obtained by setting the probability of same language to 1.0 for same-language trials and to 0.0 for cross-language trials. For the remaining experiments, we use a probability of same language of 0.5. 

\subsection{Method Comparison}

We now compare the performance of the four methods on all test sets from Mixer and LASRS divided by microphone type using the two training sets: FULL and SINGLE-LAN. 

\begin{figure}[!t]
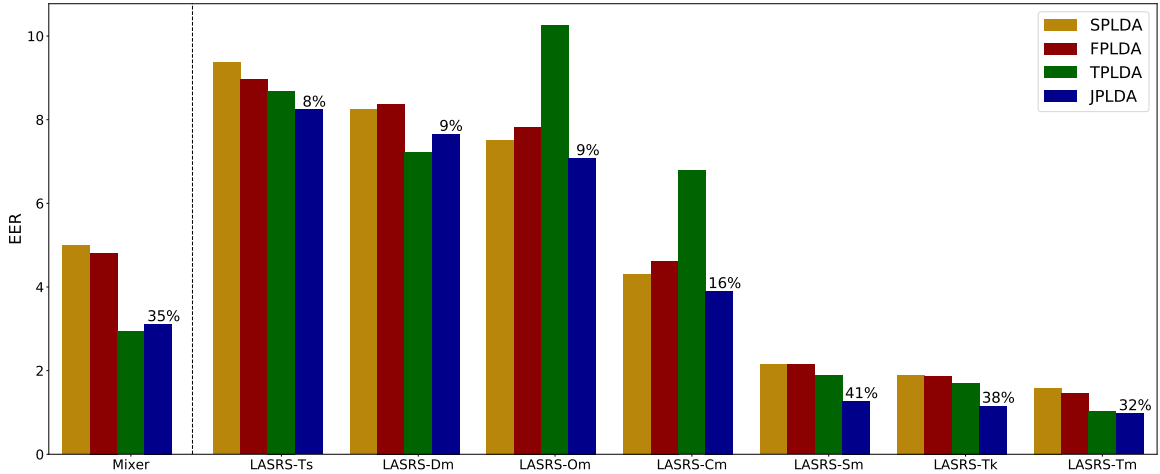
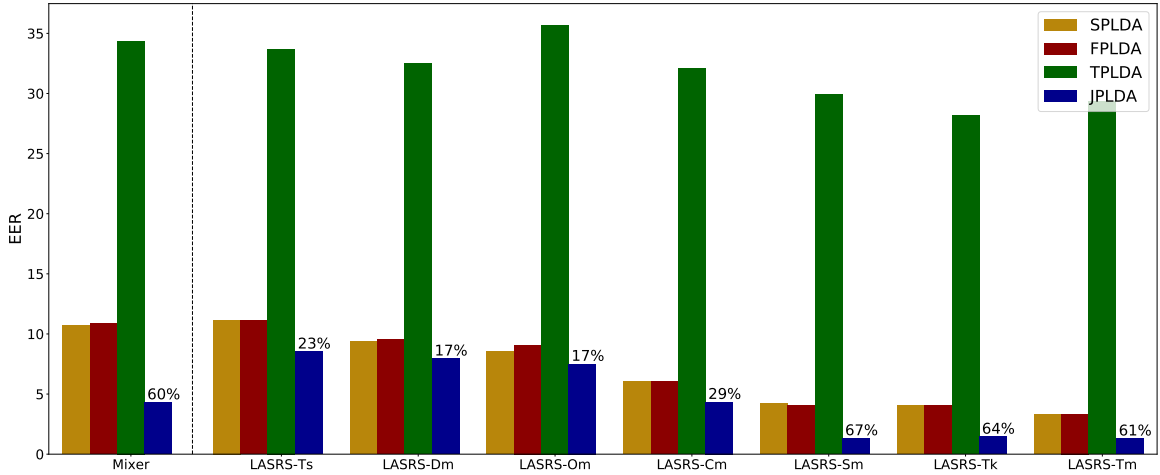

\centering
\subfloat[Training data: FULL]{\includegraphics[width=1.0\columnwidth]{{{figs/METHODS}}}}
\hfil
\subfloat[Training data: SINGLE-LANG]{\includegraphics[width=1.0\columnwidth]{{{figs/METHODS_1langxspkr}}}}
\caption{Comparison of performance for four PLDA methods on all test sets using both training sets, FULL and SINGLE-LAN. The numbers on top of the JPLDA bars show the relative gain of JPLDA relative to FPLDA.}
\label{fig:methods}
\end{figure}

The top plot in Figure \ref{fig:methods} shows that FPLDA gives slightly better performance than SPLDA for some channels (mostly the telephone ones) when the FULL training data is used. For this reason, for the remaining experiments in this paper, we use FPLDA as the baseline. 

Comparing the two methods that consider language labels during training, TPLDA and JPLDA, on the top plot in Figure \ref{fig:methods}, we see that they both give significant gains over the baselines on Mixer data, where the channel is matched to the majority of the training data's channel. In this case, both approaches succeed in mitigating the effect of language variability. On the other hand, when the channel is not exactly the same as the one observed most in training, TPLDA fails to generalize, leading to consistently worse performance than JPLDA. This is reasonable: while alternative microphone data is observed for the English training data, only telephone data is observed for the non-Englishd data. This implies that the PLDA mixture corresponding to non-English data in TPLDA was only learned with telephone data, resulting in the poorer performance observed on some of the LASRS channels. On the other hand, JPLDA can leverage the information about alternative microphones learned from English data for all languages, since the matrix that models this variability is shared across languages.

In the bottom plot in Figure \ref{fig:methods}, we see that when only a single language from each speaker is available for training (that is, the within speaker variation due to language is not observed in training), TPLDA leads to a large degradation over both baselines. Note that, as far as we know, TPLDA had not been tested under this challenging scenario. Rather, it was tested using training data where each class of interest (e.g., a face) was seen under all possible conditions (front and profile) \citep{peng2012}. When each class is seen under a single condition, the TPLDA model basically degenerates to separate (untied) PLDA models, each learned on the data from its own condition. This implies that the resulting mixture will be unable to  model the cross-language variability, which results in extremely degraded performance on the cross-language trials. Indeed, our results indicate that the same-language trials get reasonable TPLDA performance (results not shown here), it is the degradation on the cross-language trials that affects the overall performance as observed in the plot.

Finally, focusing on JPLDA, we see that significant gains are observed compared to both baselines using both training sets, with larger relative gains when the training data contains only a single language per speaker, in which case we find gains from 17\% of up to 67\% relative to the FPLDA baseline.

\subsection{Training Data Comparison}
\label{res:train}

Finally, in this section we compare the FPLDA baseline and JPLDA using the two training sets defined in Section \ref{sec:training_data} and their subsets, where we discard telephone samples from speakers that only have English samples in an attempt to achieve a better balance between English and non-English samples and telephone and microphone samples. 

\begin{figure}[!t]
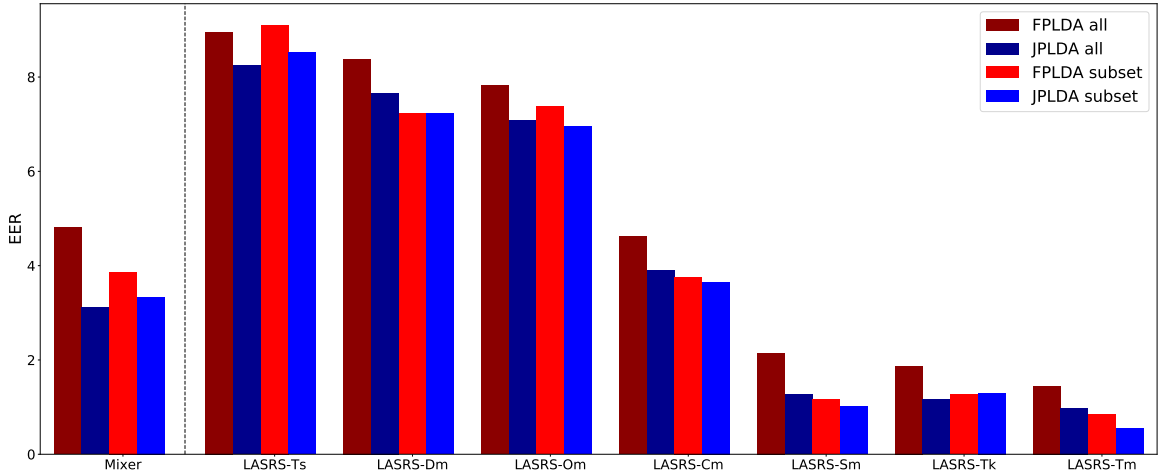
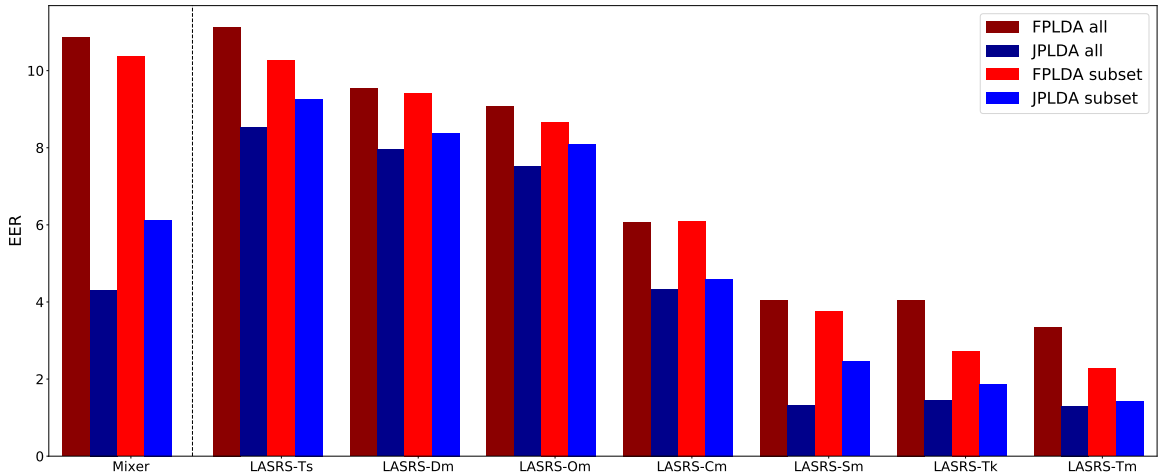

\centering
\subfloat[Training data: FULL]{\includegraphics[width=\columnwidth]{{{figs/FINAL}}}}
\hfil
\subfloat[Training data: SINGLE-LANG]{\includegraphics[width=\columnwidth]{{{figs/FINAL_1langxspkr}}}}
\caption{Comparison of performance for FPLDA and JPLDA on all test sets using the two training sets, FULL and SINGLE-LAN. For each case, we compare using the full training set and a subset where we discard telephone samples from speakers that only have English samples in the FULL training set.}
\label{fig:final}
\end{figure}

Figure \ref{fig:final} shows that, for FPLDA, using the subset is significantly better than using the full training set for both training conditions, FULL and SINGLE-LANG, for most test conditions. That is, FPLDA benefits from having a more balanced distribution of conditions within the training data. This is because, in standard PLDA, the samples from all speakers are assumed to follow the same distribution, regardless of whether these samples are all in English, or both in English and some other language. Hence, if a large proportion of speakers only have English samples, the parameters in the PLDA model will be mostly determined by what is optimal for these speakers, degrading the performance on non-English and cross-language trials.

On the other hand, JPLDA does not seem to require subsetting the data\footnote{Note that the EER on the better performing test sets (LASRS-Sm, LASRS-Tk and LASRS-Tm) corresponds to very few misses, making that metric somewhat unreliable on those  sets. However, DET curves and Cllrs shown later in the section complement the EER results, supporting the overall conclusions made based on EERs.}. In fact, for the FULL training condition, JPLDA leads to similar or better performance (using either the full training set or the subset) than FPLDA using the subset. For the SINGLE-LANG condition, the advantage of JPLDA over FPLDA is much larger than for the FULL training set, consistently showing significant gains over the best FPLDA result. 
Further, for this training condition we see a consistent trend showing that JPLDA benefits from using the full training set, which indicates that, contrary to PLDA, JPLDA can handle the imbalance in the full set of data, successfully leveraging the additional samples missing from the subset. 

\begin{figure}[!t]
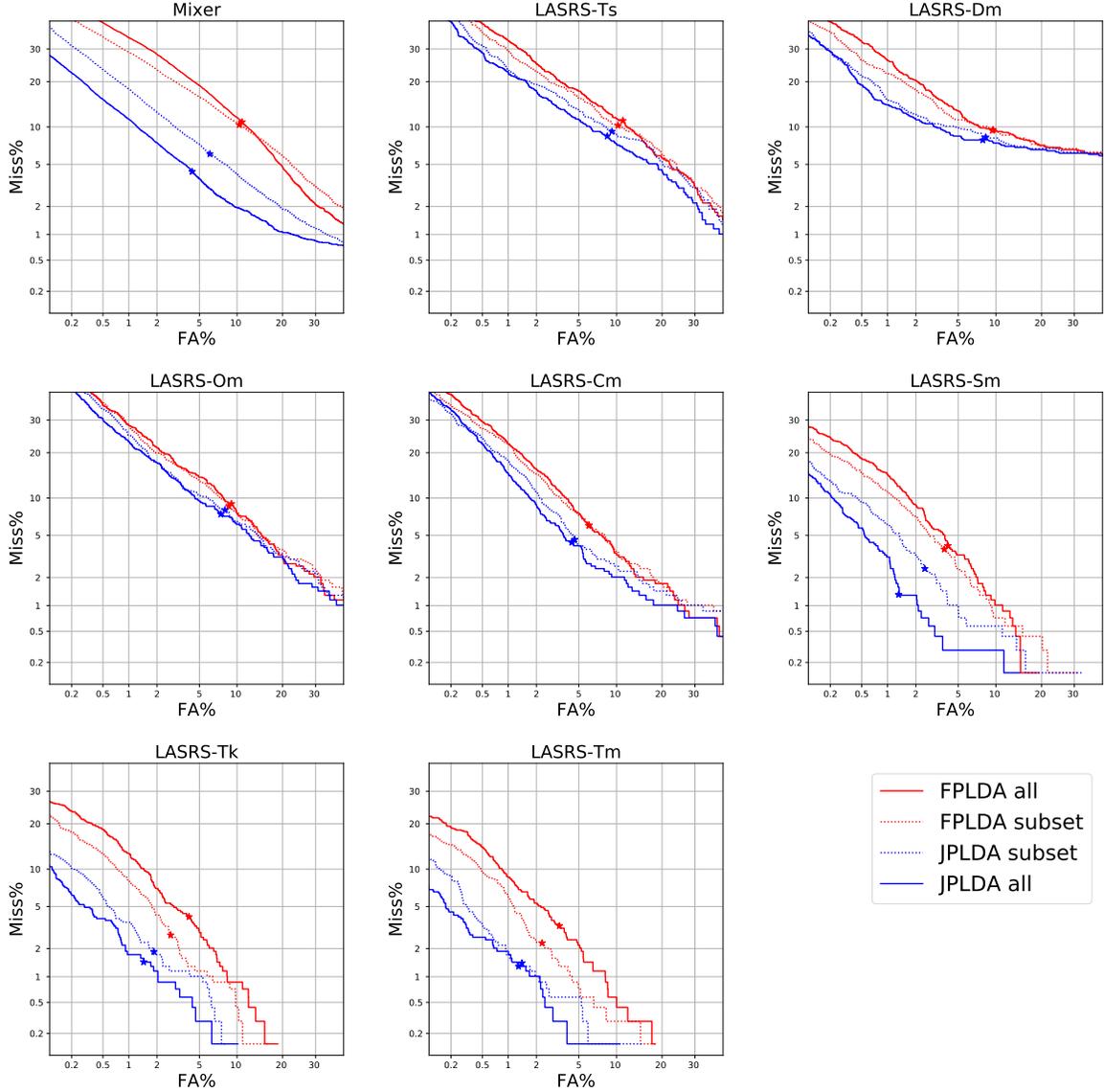

\centering
\subfloat{\includegraphics[width=0.34\columnwidth]{{{figs/dets.single_lang/mixer}}}}
\subfloat{\includegraphics[width=0.34\columnwidth]{{{figs/dets.single_lang/lasrs.Ts}}}}
\subfloat{\includegraphics[width=0.34\columnwidth]{{{figs/dets.single_lang/lasrs.Dm}}}}
\vspace{-0.5cm}
\hfil
\subfloat{\includegraphics[width=0.34\columnwidth]{{{figs/dets.single_lang/lasrs.Om}}}}
\subfloat{\includegraphics[width=0.34\columnwidth]{{{figs/dets.single_lang/lasrs.Cm}}}}
\subfloat{\includegraphics[width=0.34\columnwidth]{{{figs/dets.single_lang/lasrs.Sm}}}}
\vspace{-0.5cm}
\hfil
\subfloat{\includegraphics[width=0.34\columnwidth]{{{figs/dets.single_lang/lasrs.Tk}}}}
\subfloat{\includegraphics[width=0.34\columnwidth]{{{figs/dets.single_lang/lasrs.Tm}}}}
\subfloat{\includegraphics[width=0.34\columnwidth]{{{figs/dets.single_lang/legend}}}}
\caption{DET curves for FPLDA and JPLDA on all test sets using the SINGLE-LAN training set and its subset. The marker over each curve corresponds to the EER point for that system.}
\label{fig:dets}
\end{figure}

To complement the EER results in the bar plots, Figure \ref{fig:dets} shows the DET curves for all test sets. We show these curves for the more challenging training condition, SINGLE-LANG, where JPLDA gives the biggest advantage over FPLDA. The plots show that the gains are not specific to the EER operating point. Rather, JPLDA gives a significant gain over FPLDA over a very wide range of operating points corresponding to miss and false alarms rates between 0.01\% to 40\%. Further, we also see the advantage of using all the available training data rather than just the subset when using JPLDA, while the opposite is true for FPLDA, as already observed in the EER bar plots.

\begin{figure}[!t]
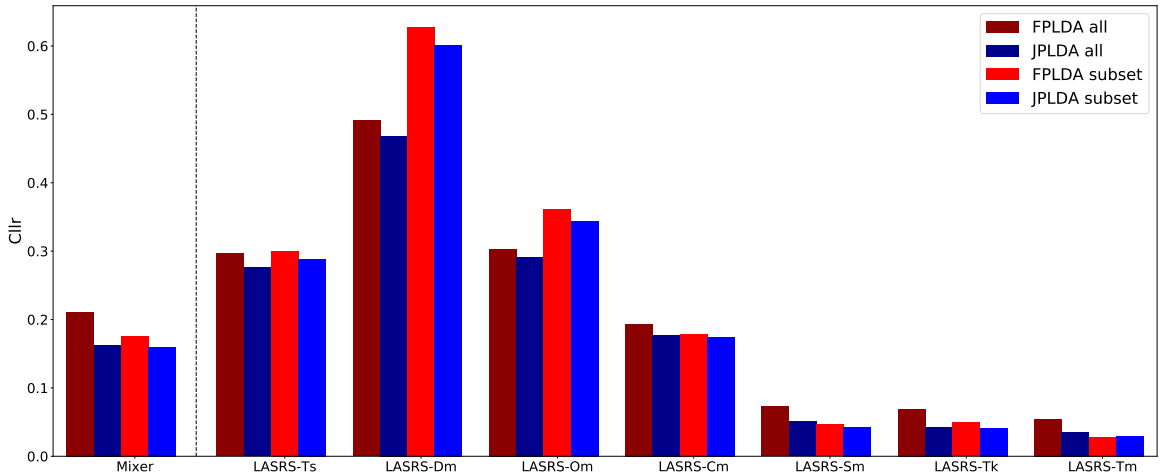
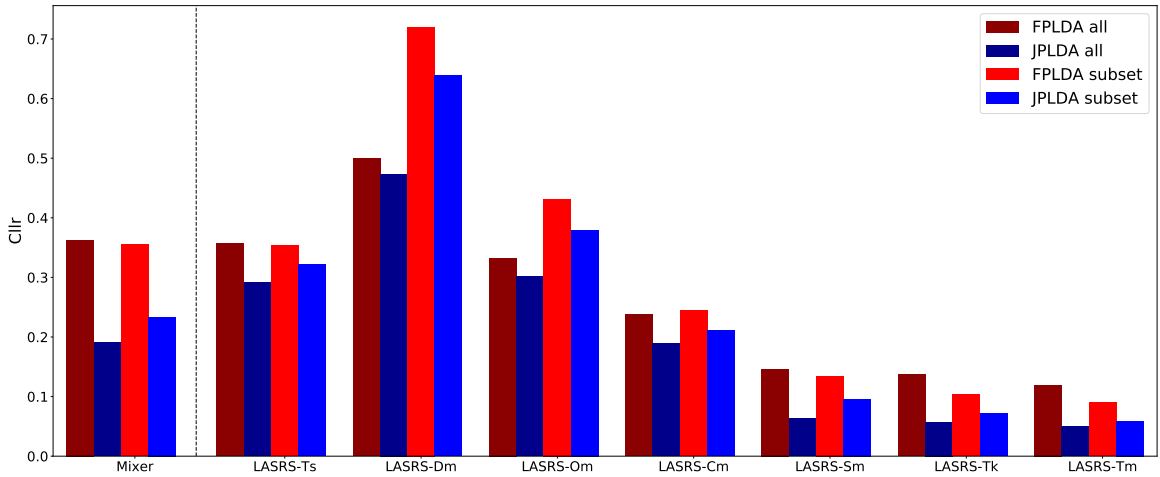

\centering
\subfloat[Training data: FULL]{\includegraphics[width=\columnwidth]{{{figs/FINAL.cllr}}}}
\hfil
\subfloat[Training data: SINGLE-LANG]{\includegraphics[width=\columnwidth]{{{figs/FINAL_1langxspkr.cllr}}}}
\caption{Same as Figure \ref{fig:final} but for Cllr instead of EER.}
\label{fig:final_cllr}
\end{figure}

Because both EER and DET curves are insensitive to calibration issues, we also show Cllr performance for all test sets on both training conditions in Figure \ref{fig:final_cllr}. The Cllr is computed after calibrating the scores using cross-validation: the scores for each test set are divided in two splits by speaker, and the data from one split is used to train a calibration model for the data in the other split. Finally, the calibrated scores from both splits are collected, and the Cllr is computed. Note that the merged scores from both splits have approximately half of the impostor samples compared to the original sets, since the trials between speakers from different splits are not included. The figure shows that JPLDA outperforms FPLDA in terms of Cllr, more so when the training data contains a single language per speaker. These observations agree with those obtained from the EER plots.

A trend that is changed in Cllr results compared to EER results is the performance when using the selected subset of the training data. For FPLDA, the subset gives inconsistent Cllr results, while for EER, using the subset was consistently better than using the complete set. That is, FPLDA requires a different selection of training data depending on the metric to be optimized. This further highlights the advantage provided by JPLDA, for which the selection of training data is not dependent on the metric.

\begin{figure}[!t]
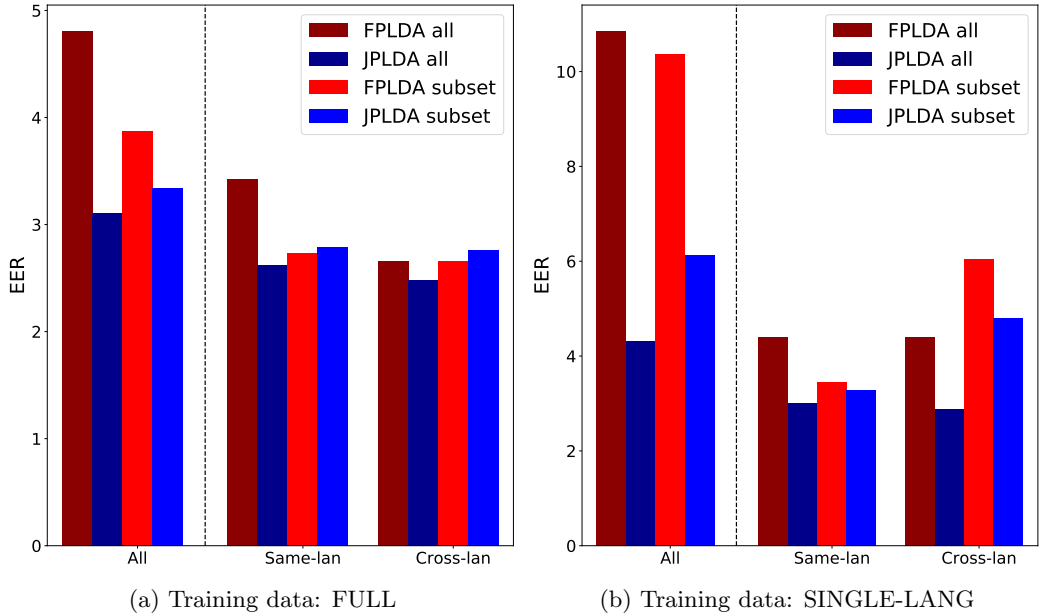

\centering
\subfloat[Training data: FULL]{\includegraphics[width=0.45\columnwidth]{{{figs/FINAL_same_vs_cross}}}}
\subfloat[Training data: SINGLE-LANG]{\includegraphics[width=0.45\columnwidth]{{{figs/FINAL_1langxspkr_same_vs_cross}}}}
\caption{Comparison of performance for FPLDA and JPLDA on the Mixer test set on all trials as well as on same-language and cross-language subsets, using the two training sets, FULL and SINGLE-LAN, and their subsets.}
\label{fig:same_cross}
\end{figure}

Finally, Figure \ref{fig:same_cross} shows EER results on Mixer test data using the two training sets and their subsets for all trials (as in previous bar plots) as well as for same-language and cross-language trials. The performance on all trials is the same as in Figure \ref{fig:final}. These plots show that: (1) Both same-language and cross-language trials benefit from using JPLDA, particularly for the SINGLE-LAN training conditions. (2) The JPLDA benefit from using the complete training sets holds for both same-language and cross-language subsets of trials. (3) The FPLDA benefit from using the subset only holds on the same-language trials; cross-language trial performance is degraded or unchanged by subsetting the training data. And (4) the relative gain from JPLDA is larger once same-language and cross-language trials are pooled together. This last observation indicates that JPLDA is not only improving discrimination for each type of trial (same-language and cross-language), but it is also aligning the distributions of these two types of trials such that when they are pooled together, the relative gain from using JPLDA is emphasized.

\section{Conclusions}

We have proposed a generalization of PLDA where within-class variability factors are no longer considered independent across samples. The method assumes that the identity of a nuisance condition is known during training and ties the latent variable corresponding to the within-class variability across all samples with the same nuisance condition label. During scoring, a likelihood ratio is computed as for standard PLDA by marginalizing over the nuisance condition. Hence, the identity of the nuisance condition can be unknown during testing.

We show results on a multilingual speaker recognition task comparing the proposed method with two types of standard PLDA models as well as to a tied PLDA model where the nuisance condition is used to determine the component in a mixture of PLDA models. Our results show that large relative gains are obtained from using JPLDA when the training data contains few or no speakers with data in more than one language. That is, the JPLDA model is able to extrapolate the effect of language from a small proportion or even zero training speakers with data from more than one language. Standard PLDA models are only able to mitigate the effect of language when exposed to a significant proportion of training speakers with data in more than one language. 

The proposed JPLDA method can be used for any task for which standard PLDA is used whenever a discrete nuisance condition is known during training. Examples include speaker recognition using channel, speaking style or language labels, among others, as the sample-dependent nuisance condition, and face recognition using pose as sample-dependent nuisance condition. The strength of JPLDA lies in its ability to extrapolate the effect that the nuisance condition has on the samples based on few or even no classes (speakers or faces) seen under several nuisance conditions. 

The proposed approach introduces the additional requirement with respect to the original PLDA approach that the identity of the nuisance condition be known during training. In future work, we will explore the possibility of automatically detecting the nuisance conditions, using classifiers trained on data for which the factors are known or using clustering with distance metrics designed to reflect the nuisance of interest. Finally, an interesting generalization of the proposed approach would be to allow for more than one sample-dependent nuisance condition. These are directions we plan to explore in the near future.

\acks{This material is based upon work supported partly by the Defense Advanced Research  Projects  Agency  (DARPA)  under  Contract  No.   HR0011-15-C-0037. The views, opinions, and/or findings contained in this article are those of the authors and should not be interpreted as representing the official views or policies of the Department of Defense or the U.S. Government.  Distribution Statement A: Approved for Public Release, Distribution Unlimited.}

\bibliography{all-short}

\end{document}